\documentclass{article}

\usepackage[preprint]{neurips_2026}

\usepackage[utf8]{inputenc} %
\usepackage[T1]{fontenc}    %
\usepackage{hyperref}       %
\usepackage{url}            %
\usepackage{booktabs}       %
\usepackage{amsfonts}       %
\usepackage{nicefrac}       %
\usepackage{microtype}      %
\usepackage{xcolor}         %

\usepackage{amsmath}
\usepackage{amssymb}
\usepackage{bbm}
\usepackage{enumitem}
\usepackage{graphicx}
\usepackage{wrapfig}
\usepackage{subcaption}
\usepackage[most]{tcolorbox}
\usepackage{xcolor}
\definecolor{promptbg}{HTML}{F7F7F7}
\definecolor{promptframe}{HTML}{D9D9D9}
\newtcolorbox{promptbox}[2][]{
    enhanced,
    breakable,
    colback=promptbg,
    colframe=promptframe,
    coltitle=black,
    fonttitle=\bfseries,
    title={#2},
    boxrule=0.5pt,
    arc=2pt,
    left=6pt,
    right=6pt,
    top=6pt,
    bottom=6pt,
    fontupper=\ttfamily\small,
    #1
}
\newcount\Comments  %
\definecolor{darkgreen}{rgb}{0,0.5,0}
\definecolor{darkred}{rgb}{0.7,0,0}
\definecolor{teal}{rgb}{0.1,0.6,0.7}
\definecolor{blue}{rgb}{0.0,0.1,0.9}
\definecolor{orange}{rgb}{1.,0.7,0.0}
\definecolor{palegreen}{rgb}{0.7,0.7,0.0}
\definecolor{lightblue}{rgb}{0.70, 0.80, 0.89}
\definecolor{violet}{rgb}{0.50, 0.16, 0.88}
\definecolor{babyblue}{rgb}{0.00, 0.88, 0.88}
\definecolor{electricpurple}{rgb}{0.75, 0.0, 1.0}

\newcommand{\kibitz}[2]{\ifnum\Comments=1{{\textcolor{#1}{\textsf{\footnotesize [#2]}}}}\fi}

\newcommand{\eat}[1]{}

\newcommand{\autodv}{AutoDiscovery}
\newcommand{\ver}{\mathcal{V}_D}

\DeclareRobustCommand{\ai2}{%
  \begingroup\normalfont
  \includegraphics[height=1.2\fontcharht\font`\M]{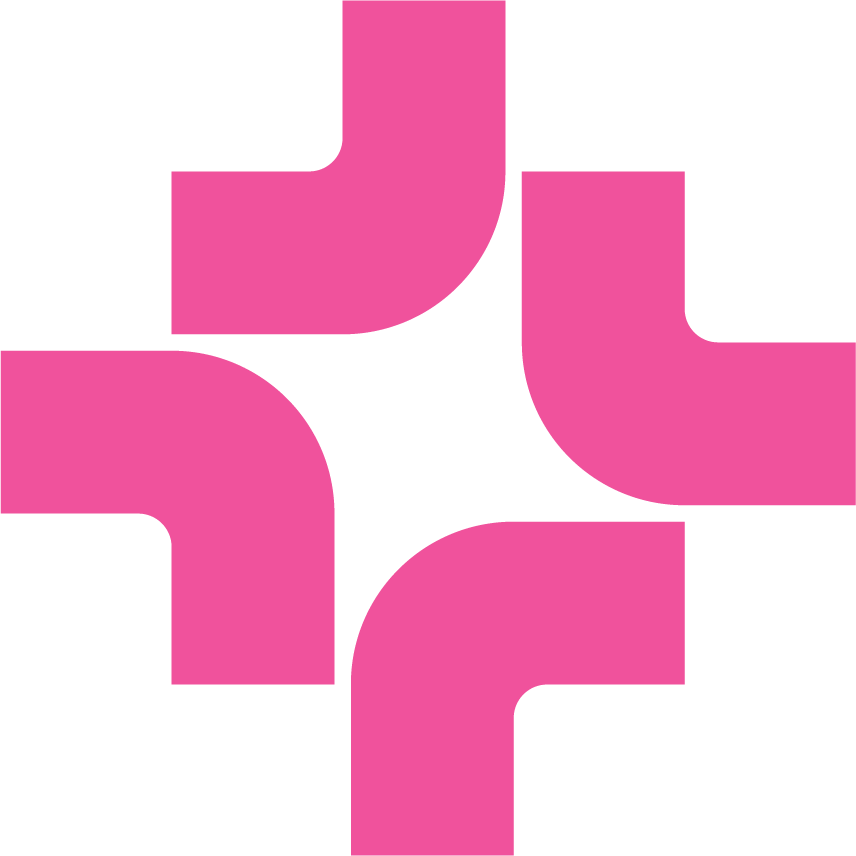}%
  \endgroup
}

\title{Evidence-Informed LLM Beliefs\\ for Continual Scientific Discovery}

\author{
    Dhruv Agarwal$^{\textcolor[HTML]{f0529c}{\alpha\beta}}$ ~
    Reece Adamson$^{\textcolor[HTML]{f0529c}{\beta}}$ ~
    Andrew McCallum$^{\textcolor[HTML]{f0529c}{\alpha}}$ ~
    \\\\
    \textbf{
    Peter Clark$^{\textcolor[HTML]{f0529c}{\beta}}$ ~
    Ashish Sabharwal$^{\textcolor[HTML]{f0529c}{\beta}}$ ~
    Bodhisattwa Prasad Majumder$^{\textcolor[HTML]{f0529c}{\beta}}$
    } \\\\
    $^{\textcolor[HTML]{f0529c}{\alpha}}$University of Massachusetts Amherst ~ $^{\textcolor[HTML]{f0529c}{\beta}}$Allen Institute for AI \\
    \texttt{dagarwal@cs.umass.edu, bodhisattwam@allenai.org}\\
    \\ \ai2~\url{https://github.com/allenai/autodiscovery-continual-beliefs}
}

\begin{document}

\maketitle

\begin{abstract}
Open-ended scientific discovery with large language models (LLMs) increasingly operates as a long-horizon loop of hypothesis search and verification, where a reward signal guides which hypotheses to test next. 
A notable recent example is AutoDiscovery \citep{agarwal2025autodiscovery}, which uses ``Bayesian surprise''---the belief shift an LLM undergoes after observing evidence for a hypothesis---as both a discovery metric and a reward for search.
We first observe that \autodv{} treats surprisal as a \emph{static} quantity, while surprisal in human reasoning is \emph{non-stationary}---it is defined relative to beliefs that evolve with experience, a prerequisite for continual scientific discovery.
We address this mismatch with \emph{evidence-informed LLM beliefs}: priors updated with evidence from previous hypotheses to compute \emph{non-stationary surprisal} for new hypotheses.
We compare in-context belief-updating mechanisms and find that embedding-based retrieval-augmented generation over prior discoveries best anticipates eventual posteriors, identifying 37.5\% of static surprisals as spurious.
We then modify search to avoid these spurious rewards and prioritize hypotheses that remain surprising under non-stationary beliefs.
Concretely, we introduce two complementary changes to the original search procedure: belief-update filtering and diversity maximization.
Across five discovery domains, our method increases accumulated non-stationary surprisal by 30.62\% on average compared to the original search procedure, demonstrating that continual scientific discovery with LLMs requires not only better belief measurement but also search procedures that avoid redundancy and encourage diversity.
\end{abstract}

\section{Introduction}
\label{sec:introduction}

Scientific discovery is inherently \emph{continual}––each observation alters what should be considered novel, informative, or surprising going forward. 
This view is central to formal accounts of scientific inquiry, where evidence goes beyond isolated hypothesis verification to also revise the epistemic state of a discovery agent, 
thus changing the space of plausible future hypotheses \citep{alchourron1985logic,gardenfors1988knowledge,martin1997scientific}. 
In Bayesian accounts of scientific reasoning, this revision is captured by conditioning, where evidence transforms a prior set of beliefs into a posterior set of beliefs, and subsequent inference proceeds from the updated posterior rather than the original prior \citep{earman1992bayes,howson2006scientific}. 
Similarly, computational accounts of how scientific theories evolve emphasize that hypotheses are accepted or rejected not independently, but through their coherence with accumulated explanations and observations \citep{thagard1989explanatory,thagard1992conceptual,jansen2026generatingliteraturedrivenscientifictheories}. 
Thus, in a continual discovery setting, the belief state should be \emph{non-stationary} and evolve in light of previous results. For instance, when assessing novelty, once a finding is observed, other hypotheses entailed or strongly suggested by this finding should be considered \emph{less} novel.

\begin{figure}[t]
    \centering
    \includegraphics[width=\textwidth]{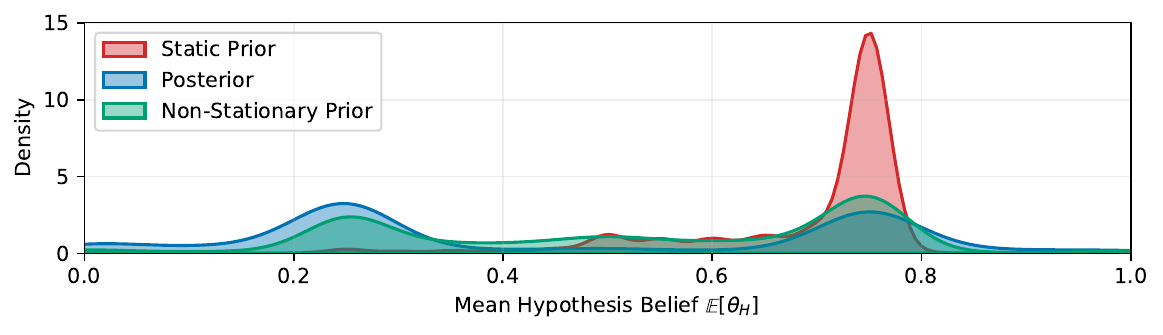}
    \caption{
        \textbf{Static vs. non-stationary beliefs.}
        Belief distributions for 7500 hypotheses found by \autodv{} across five discovery domains. \autodv{} uses \emph{static} beliefs to score hypotheses using an unchanged LLM prior, which causes discoveries already implied by past evidence from search to spuriously appear novel.
        Evidence-informed LLM priors move the reference belief state towards the eventual posterior and allow for \emph{non-stationary} surprisal to be computed instead, lowering surprisal count by 37.43\% compared to what was found by the original search. 
    }
    \label{fig:hyp-belief-dist}
\end{figure}

Automated scientific discovery with large language models (LLMs) has begun showing promise in such continual settings, alternating between proposing hypotheses (``search'') and evaluating their support (``verification'') over long horizons \citep{aiscientist, yamada2025aiscientistv2workshoplevelautomated, aicoscientist}.
A notable recent example of such a system is AutoDiscovery \citep{agarwal2025autodiscovery}, which uses ``Bayesian surprise'' \citep{itti2005bayesian} as a reward for hypothesis search---given a hypothesis $H$, the system elicits a prior belief $P(\theta_H)$ about the support for $H$ according to an LLM, evaluates $H$ using a verification procedure $\ver$ over available data $D$, then elicits a posterior belief $P(\theta_H \mid \ver)$ to compute a reward according to the resulting belief shift. This reward is then used to steer search towards candidate hypotheses that are likely to surprise the LLM as a proxy signal for discovery.

However, we observe that in both the surprisal reward and evaluation metric, AutoDiscovery treats the LLM prior as \emph{static}---each hypothesis is evaluated against the model's parametric beliefs in isolation, without conditioning on discoveries made earlier in the search trajectory. 
This static treatment creates a mismatch between LLM-driven discovery and human scientific reasoning---as an evaluation metric, static surprisal overestimates progress by counting derivative hypotheses as successful discoveries; as a search reward, it wastes budget by repeatedly steering the agent toward saturated regions of the hypothesis space. 
Through a human expert study in neuroscience and social science, we find that 29.54\% and 25\% of hypotheses found surprising by \autodv{} in each domain, respectively, are substantially implied by its previous discoveries rather than being genuinely novel.

Akin to Bayesian models of cognition that treat human learning as probabilistic inference over hypotheses, where observations update beliefs that support further prediction and exploration \citep{tenenbaum2006theory,griffiths2008bayesian,tenenbaum2011grow}, we argue that LLM-driven discovery must instead utilize \emph{non-stationary} surprisal using prior beliefs that are continually updated with evidence from previous hypotheses.
We compare different mechanisms for updating beliefs using in-context learning, including a structured memory baseline \citep{chhikara2025mem0}, and find that simple top-$k$ embedding-based retrieval-augmented generation (RAG) best integrates prior evidence to anticipate the eventual posterior, lowering total surprisal by 37.43\%\footnote{Hypotheses where evidence-informed priors match the posteriors after verification.} on average. As shown in Figure~\ref{fig:hyp-belief-dist}, the distribution of non-stationary prior beliefs is able to move closer to the experimental posterior. 

We then investigate whether the search budget lost to spurious surprisal can be reallocated to finding hypotheses that remain surprising under non-stationary beliefs. 
To do so, we present evidence-informed search in \autodv{} introducing two complementary mechanisms: (a) belief-update filtering, to sample hypotheses that cannot be derived in-context using prior discoveries, and (b) diversity maximization, which explicitly prioritizes exploration by selecting hypotheses with the least embedding-based similarity to previous discoveries.
Across five discovery domains \citep{majumder2025discoverybench,gu2024bladebenchmarkinglanguagemodel}, these modifications increase accumulated non-stationary surprisal by 30.62\% on average over the original search procedure, showing that continual discovery requires search policies that explicitly manage redundancy, belief saturation, and diversity.

In summary, our contributions are:
\begin{itemize}[leftmargin=0.5cm]
    \item We identify a failure mode of \emph{static surprisal} in continual LLM-driven discovery: hypotheses are scored against unchanged priors despite accumulating evidence during search. We validate this failure mode with human experts in neuroscience and social science, showing that 29.54\% and 25\% of hypotheses considered surprising using static beliefs, respectively, are derivable from prior discoveries.

    \item We introduce \emph{non-stationary surprisal}, which evaluates hypotheses relative to LLM beliefs updated with evidence from previously tested hypotheses. We compare ICL-based belief-updating mechanisms and find that embedding-based RAG best anticipates posterior beliefs, lowering total surprisal count by 37.5\% of the number of static surprisals.

    \item We present evidence-informed search for \autodv{} with belief-update filtering and diversity maximization, yielding a 30.62\% average increase in accumulated non-stationary surprisal across five discovery domains.
\end{itemize}

\section{Background}
\label{sec:background}

\paragraph{Data-driven discovery.} Following \citet{majumder2025discoverybench}, we define a \emph{data-driven hypothesis} $H \in \mathcal{H}$ as a natural language statement. Given a dataset $D$, the truth value of $H$ is determined by a verification procedure $\mathcal{V}_D:\mathcal{H} \to \{\mathrm{supported}, \mathrm{unsupported}\}$, where $\mathcal{V}_D$ may be any executable Python program. 

\paragraph{AutoDiscovery.} Our focus in this work is on a recent open-ended discovery system called AutoDiscovery \citep{agarwal2025autodiscovery}, where, given a dataset $D$ and a search budget, a discovery agent iteratively searches for promising hypotheses driven by ``Bayesian surprise'' as a reward using MCTS with Progressive Widening\citep{coulom2006efficient, couetoux2011continuous}. Here, we use a variation of MCTS, UCB1 Recursive, which greedily selects the next state for expansion among a node and its children in a recursive manner until either the parent node or a leaf is selected.
Following \citet{agarwal2025autodiscovery}, we measure Bayesian surprise $BS$ for a hypothesis by the change in its expected belief before and after results from its verification procedure are observed.
To elicit beliefs, multiple responses are sampled from an LLM based on five categories: \emph{``definitely false''}, \emph{``maybe false''}, \emph{``uncertain''}, \emph{``maybe true''}, and \emph{``definitely true''}, which are then mapped to pseudo-Bernoulli counts by assigning scores of $0$, $0.25$, $0.5$, $0.75$, and $1.0$. These counts can then converted into a Beta distribution using the Beta-Bernoulli conjugacy. For simplicity and interpretability, we evaluate belief change using empirical surprisal, defined as the difference between the posterior and prior empirical means. Each empirical mean is simply the average numerical score assigned to the sampled categorical responses, so the resulting quantity directly reflects the shift in the model’s elicited belief on the $[0,1]$ scale.

\section{From Static to Non-Stationary Surprisal}
\label{sec:static-to-non-stationary}

Consider an agent at timestep $t$ of the continual discovery process in \autodv{} after observing hypotheses $\{H^{(i)}\}_{i=1}^{t-1}$ and their verification outcomes $\{\ver^{(i)}\}_{i=1}^{t-1}$. 
To evaluate a new hypothesis $H^{(t)}$, 
a prior belief $P(\theta_{H^{(t)}})$ is elicited from an LLM, which is then standardly used when measuring surprisal with respect to a posterior belief $P(\theta_{H^{(t)}} \mid \ver^{(t)})$.
However, previous work elicits the prior independently for each new hypothesis from the same \emph{static}\footnote{Encoded within LLM parameters during various stages of model training.} belief distribution.

\paragraph{Failure mode with static priors.}
We first observe that using static beliefs in continual discovery results in a systematic failure mode. 
Hypotheses that are implied---partially or completely---by earlier discoveries may still elicit a significant prior-to-posterior shift, thereby incorrectly registering as surprising.
Consequently, the discovery agent may repeatedly propose derivative hypotheses with low information gain or even ``rediscover'' previous hypotheses, spuriously inflating reward for saturated regions of the search space (i.e., nodes in the search tree), which results in a wastage of search budget. Thus, the relevant prior for evaluating a new hypothesis $H^{(t)}$ is not a \emph{static}, unconditional belief $P(\theta_{H^{(t)}})$, but a belief state updated with previous discoveries as evidence,
\[
    P\!\left( \label{eq:non-stationary-prior}
        \theta_{H^{(t)}} 
        \mid 
        \{(H^{(i)}, \ver^{(i)})\}_{i=1}^{t-1}
    \right).
\]
We define the above expression as a \textbf{non-stationary prior}, which allows us to formally define \textbf{non-stationary surprisal} as,
\[
    S_\mathrm{NS}(H^{(t)},\ver) := \mathbbm{1} \left[ \left| \label{eq:non-stationary-surprisal}
    \mathbb{E}_{P(\theta_{H^{(t)}} \mid \ver)}[\theta_{H^{(t)}}]
    - \mathbb{E}_{P(\theta_{H^{(t)}} \mid \{(H^{(i)}, \ver^{(i)})\}_{i=1}^{t-1})}[\theta_{H^{(t)}}]
    \right| \ge \tau \right ],
\]
where $\tau$ is a threshold used to determine if a belief shift should be marked as a surprisal or not\footnote{We set this value to $0.3$ throughout this paper.}. This non-stationary view is consistent with formal and Bayesian accounts of scientific reasoning, where evidence revises an agent's epistemic state and changes which future hypotheses should be considered plausible, informative, or surprising \citep{alchourron1985logic,gardenfors1988knowledge,martin1997scientific,earman1992bayes,howson2006scientific,thagard1989explanatory,thagard1992conceptual,jansen2026generatingliteraturedrivenscientifictheories,tenenbaum2006theory,griffiths2008bayesian,tenenbaum2011grow}.

\paragraph{Non-stationarity in human beliefs.}
We validate this premise with human experts in neuroscience and social science on hypotheses generated by \autodv{}. 
In each domain, we ask experts to annotate their prior belief for 50 hypotheses when the top-5 most similar previous hypotheses and their verification outcomes are made available to them. 
First, we find that experts indeed \emph{update} their beliefs from the stated prior based on past discoveries in 48.07\% and 52\% cases, respectively.
Further, in 29.54\% (of 44) and 25\% (of 32) hypotheses considered surprising by \autodv{}, experts agree that evidence from past discoveries would be informative, hence may not be truly surprising discovery. 
This confirms that the \emph{static} treatment of beliefs in \autodv{} may be misaligned with human judgements in continual discovery. We provide additional details on the study in Appendix~\ref{app:human-study}.

\subsection{Continually Updating LLM Priors}
\subsubsection{In-Context Memory}
A natural way to incorporate evidence from previous discoveries is via in-context learning\footnote{Our study focuses on nonparametric methods to capture the discovery context, given the use of proprietary LLMs with only black-box access. Investigating parametric methods represents an important future direction.} \citep{brown2020language}. Indeed, recent work shows that accumulating interaction history in-context can substantially change the stated beliefs and downstream behavior in LLMs \citep{geng2025accumulating}, though this malleability is often framed as a reliability concern. 
Instead in this work, we use it constructively as a trajectory-specific nonparametric memory to continually update beliefs as relevant evidence accumulates.

\paragraph{Context Representations.}
We define $\mathcal{C}(\{(H^{(i)}, \ver^{(i)})\}_{i=1}^{t-1})$ as an in-context representation of evidence from previous discoveries in the search trajectory.
The simplest approach $\mathcal{C}_\text{full}$ injects all previous discoveries and their verification outcomes in the context, providing complete information to the LLM but incurring a rapidly growing context length.
Next, we evaluate imposing some structure via top-$k$ embedding-based retrieval $\mathcal{C}_{\text{top-}k}$, which selects the $k$ previously verified hypotheses most semantically similar to $H^{(t)}$. 
Finally, we also compare against a structured long-term memory baseline, Mem0 \citep{chhikara2025mem0} ($\mathcal{C}_\text{mem0}$), which dynamically extracts, consolidates, and retrieves salient information from prior interactions jointly via text embeddings and a graph structure. 

\subsubsection{Evaluation: Reducing Surprisal Under Non-Stationary Beliefs}
Our evaluation focuses on the ability of each context representation $\mathcal{C}(\cdot)$ to reduce the total surprisal $\mathcal{T} := \sum_{t=1}^{B}S(H^{(t)}, \ver)$ accumulated in a discovery run under a given search budget $B$. When using evidence-informed priors, we replace $S$ with $S_\mathrm{NS}$ in $\mathcal{T}$ and aim to align evidence-informed prior beliefs with their eventual posteriors. 
Operationally, for each hypothesis $H^{(t)}$, we elicit three beliefs: the static prior $P(H^{(t)})$, 
the evidence-informed prior $P(H^{(t)} \mid \mathcal{C}(\{(H^{(i)}, \ver^{(i)})\}_{i=1}^{t-1}))$,  
and the posterior after verification $P(H^{(t)} \mid \ver)$. 
A belief-update mechanism may result in the surprise of a hypothesis either decreasing (identified spurious surprisal), increasing (misaligned evidence), or remaining the same. 
A method is considered performant if, in aggregate, it is able to reduce a larger number of surprisals than it causes, i.e., if $\mathcal{T}$ is kept low. 

\paragraph{Experiment setup.} We run belief elicitation on hypotheses generated by \autodv{} over 5 real-world domains from DiscoveryBench (\texttt{NLS-SES}, \texttt{Archaeology}, \texttt{Evolution Freshwater Fish}; \citet{majumder2025discoverybench}) and BLADE (\texttt{Fertility}, \texttt{Mortgage}; \citet{gu2024bladebenchmarkinglanguagemodel}) using 3 repeat runs with a search budget of $n=200$ each using GPT-5-mini as the belief model.

\begin{figure}[t]
    \centering

    \begin{subfigure}{0.48\linewidth}
        \centering
        \includegraphics[width=\linewidth]{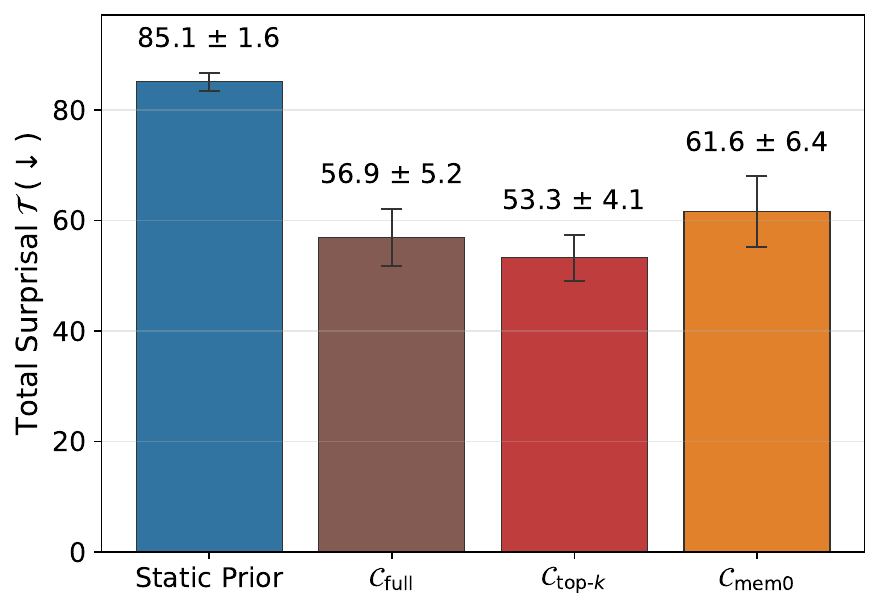}
        \caption{Comparing context representations $\mathcal{C}$}
        \label{fig:spurious-surprisal-aggregate}
    \end{subfigure}
    \hfill
    \begin{subfigure}{0.48\linewidth}
        \centering
        \includegraphics[width=\linewidth]{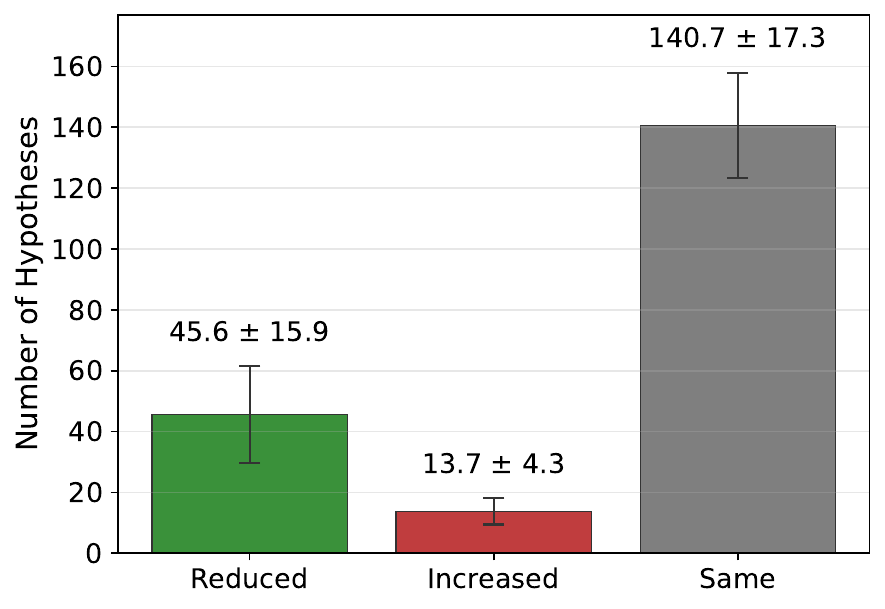}
        \caption{Effect on static surprisals with $\mathcal{C}_{\text{top-}k}$}
        \label{fig:spurious-surprisal-topk-breakdown}
    \end{subfigure}

    \caption{
    \textbf{Evidence-informed priors reduce total surprisal.}
    \textbf{(a)} Comparing context-construction methods, ICL with top-$k$ retrieval yields the lowest total surprisal $\mathcal{T}$, indicating that retrieved evidence helps align the prior with the eventual posterior.
    \textbf{(b)} Decomposing the effect of $\mathcal{C}_{\text{top-}k}$ relative to the static prior shows the proportion of surprising hypotheses from static priors that were reduced (53.5\%) or newly introduced (increased; 16\%) by evidence-informed priors. 
    Surprisal decision for 70.35\% hypotheses of the total search budget ($n=200$) remain unchanged.
    }
    \label{fig:spurious-surprisal}
\end{figure}

\paragraph{Top-$k$ representation performs best.} 
As shown in Figure~\ref{fig:spurious-surprisal}, embedding-based retrieval $\mathcal{C}_{\text{top-}k}$ (at $k=25$) is the most effective context representation for evidence-informed belief updates, showing the most reduction in $\mathcal{T}$ by 31.8 surprisals on average across repeat runs with respect to 85.1 static surprisals. In comparison, $\mathcal{C}_\text{full}$ has 28.2 fewer surprisals and $\mathcal{C}_\text{mem0}$ has 23.5.

\paragraph{Effect of $k$ and reasoning effort.} To assess the effect of the number of retrievals $k$, we show in Figure~\ref{fig:spurious-surprisal-long-horizon} a comparison of $\mathcal{C}_{\text{top-}k}$ at $k=\{1,5,25,50,\text{all}\}$ in a longer-horizon setting of $n=800$ experiments. Our results show that full context $\mathcal{C}_\text{full}$ has 6.1 percentage points greater total surprisal ($\approx 23$ hypotheses) than $\mathcal{C}_{\text{top-25}}$, while costing $8.45\times$ the number of tokens on average ($\approx 12k$ more). We also find that total surprisal is similar at $k=5$, while increasing at $k=1$ and $k=50$. Balancing cost and performance, we choose $\mathcal{C}_{\text{top-25}}$ as the context representation to compute non-stationary surprisal in the rest of the paper, and also use this as the new metric to report in \S\ref{sec:improving-search}. \ Additionally, in Figure~\ref{fig:spurious-surprisal-long-horizon-cost}, we show the cost-performance tradeoff when using different levels of LLM \texttt{reasoning effort}, and use the ``elbow'' as the point of diminishing returns to select \texttt{low} as our default setting for all subsequent evaluations.

\section{Improving Search for Continual Discovery}
\label{sec:improving-search}

The previous section shows that static surprisal overestimates discovery progress by rewarding hypotheses already implied by earlier discoveries. 
This raises a complementary optimization question: 
how can we modify the search process so as to maximize the number of hypotheses produced with high non-stationary surprisal?

\begin{wrapfigure}{l}{0.55\linewidth}
    \vspace{-0.9em}
    \centering
    \includegraphics[width=\linewidth]{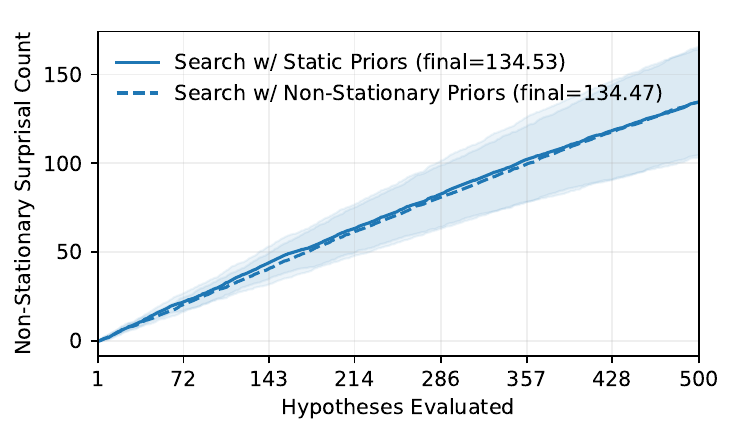}
    \vspace{-1.5em}
    \caption{
    \textbf{Discovery with static vs. non-stationary beliefs.}
        The trajectories are nearly identical, indicating that replacing the reward alone is insufficient to make the original search respond to non-stationary beliefs.
    }
    \label{fig:search-static-nonstationary}
    \vspace{-1.5em}
\end{wrapfigure}

The most direct intervention is to replace \autodv{}'s static surprisal reward with non-stationary surprisal. 
However, as shown in Figure~\ref{fig:search-static-nonstationary}, this substitution alone has no effect on discovery yield---static- and non-stationary reward runs, although exploring different hypotheses, in the end produce a nearly identical number of non-stationary surprisals. 
This suggests that changing the scalar reward is insufficient to redirect exploration toward hypotheses that remain novel under updated beliefs.

To address this, we consider ways to make the search more sensitive to evidence-informed prior, specifically via two mechanisms that directly guide the search during the hypothesis selection step.
We recall that during search in \autodv{}, once a node is selected for expansion by MCTS based on a UCB1 selection policy, the system randomly samples one of a pre-generated set of untried hypotheses to verify next.
Instead, we propose to replace this random hypothesis selection with two complementary mechanisms---\emph{belief-update filtering} and \emph{diversity maximization}.
These mechanisms help in two distinct ways: by disregarding hypotheses that are influenced strongly by some prior evidence, and by pushing the search towards semantic regions away from everything that has already been explored. 
We refer to our combination of the two mechanisms as \textit{evidence-informed search}.
Next, we discuss these mechanisms in more detail.

\paragraph{Belief-update filtering.}
First, we filter hypotheses whose static surprisal is likely to be explained well by prior discoveries, making the non-stationary surprisal low.
To this end, for each candidate hypothesis $H$, we construct an evidence-informed context using $\mathcal{C}_{\text{top-25}}(\cdot)$ and elicit an updated prior conditioned on this context.
If the absolute shift between the static prior and the evidence-informed prior exceeds a threshold $\kappa$, i.e.,
\[
    \left|
    \mathbb{E}_{P(\theta_H \mid \mathcal{C}_{\text{top-25}}(\cdot))}[\theta_H]
    -
    \mathbb{E}_{P(\theta_H)}[\theta_H]
    \right| > \kappa,
\]
then $H$ is unlikely to yield non-stationary surprisal since previous discoveries already provide enough evidence for the LLM to have a significant impact on its belief in $H$, and hence likely anticipate the posterior of $H$.
Figure~\ref{fig:belief-update-filtering-analysis} supports this criterion: as the magnitude of the shift between the static and evidence-informed priors increases, the proportion of hypotheses that remain surprising under non-stationary beliefs decreases (we set $\kappa=0.2$ based on this analysis).

We note that absolute belief shift, which can be computed without knowing the posterior, is identical to absolute surprisal shift when both the static prior and the non-stationary prior are on the same side of the posterior (i.e., both are less or both are more). In general, absolute belief shift is at least as large as the absolute surprisal shift, approaching equality as either prior gets close to the posterior. This implies that regions with relatively low belief shift given prior evidence are guaranteed to have relatively low surprisal shift as well, making them promising candidates to explore.

\paragraph{Diversity maximization.}
Second, we encourage search to explore semantically diverse regions of the hypothesis space.
We embed each candidate hypothesis and compare it to its top-$k$ most similar previously verified hypotheses.
For a candidate $H$, we define its local similarity score as the average distance to these previously verified hypotheses:
\[
    s_k(H) = \frac{1}{k}\sum_{H' \in \mathrm{topk}(H; \mathcal{H}_{<t})}
    \mathrm{sim}_\phi(H, H'),
\]
where $\phi(\cdot)$ is the embedding function, $\mathcal{H}_{<t}$ is the set of previously verified hypotheses, and $\mathrm{topk}(H; \mathcal{H}_{<t})$ denotes the $k$-nearest previous hypotheses to $H$.
We then select the candidate with the lowest $s_k(H)$, prioritizing hypotheses that are relatively far, and thus less likely to be redundant or derivable under evidence-informed beliefs.

We note that this mechanism is distinct from belief-update filtering introduced earlier; neither subsumes the other. For instance, if a selected hypothesis $H$ is relatively far from prior hypotheses, then a closely related next candidate hypothesis $H'$ (e.g., the negation of $H$) will also be relatively far (on average) from prior hypotheses including $H$---and thus desirable according to this mechanism. However, $H'$ would be strongly implied by a prior hypothesis $H$, resulting in it being filtered out due to a large belief shift.

In addition to the above modifications, we find that including an \textbf{online de-duplication} procedure that removes hypotheses that are semantically redundant with previously verified hypotheses or with other candidates proposed for the current node improves search performance when measured via both static (as in original \autodv{}) as well as non-stationary surprisal. 
We use the LLM-based hierarchical agglomerative clustering (HAC) procedure \citep{clusterllm} from \autodv{}, but apply it \emph{online} during each search iteration rather than only as a post-hoc step after the discovery run.
Candidates that cluster with an already verified hypothesis are discarded, thus preventing search from spending budget on near-duplicate past discoveries.

Note that online de-duplication is not subsumed by either of the first two mechanisms. For instance, a near-duplicate candidate hypothesis $H'$ of a prior hypothesis $H$ would not be filtered out by belief-update filtering if $H$ was not surprising to begin with---its static prior, non-stationary prior, and posterior would all be roughly equal, resulting in no or little belief shift. Similarly, as mentioned earlier, if $H$ is far on average from other prior hypotheses, then so would candidate $H'$ be. It would thus be prioritized by the diversity maximization objective (which looks at the \emph{average} distance from the top 25 closest matches). Online de-duplication, however, will correctly discard $H'$.

\subsection{Experiments}
\begin{figure}[t]
    \centering

    \begin{subfigure}[t]{\textwidth}
        \centering
        \includegraphics[width=\textwidth]{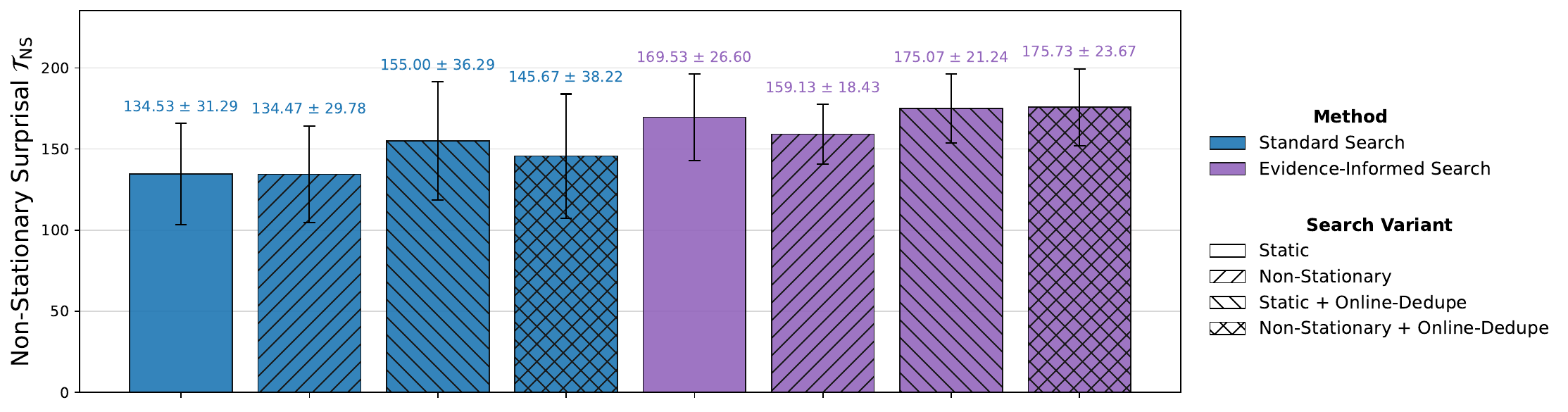}
        \caption{Comparison between variants of standard \autodv{} search and evidence-informed search.}
        \label{fig:search-results-main-panel}
    \end{subfigure}

    \vspace{0.75em}

    \begin{subfigure}[t]{0.75\textwidth}
        \centering
        \includegraphics[width=\textwidth]{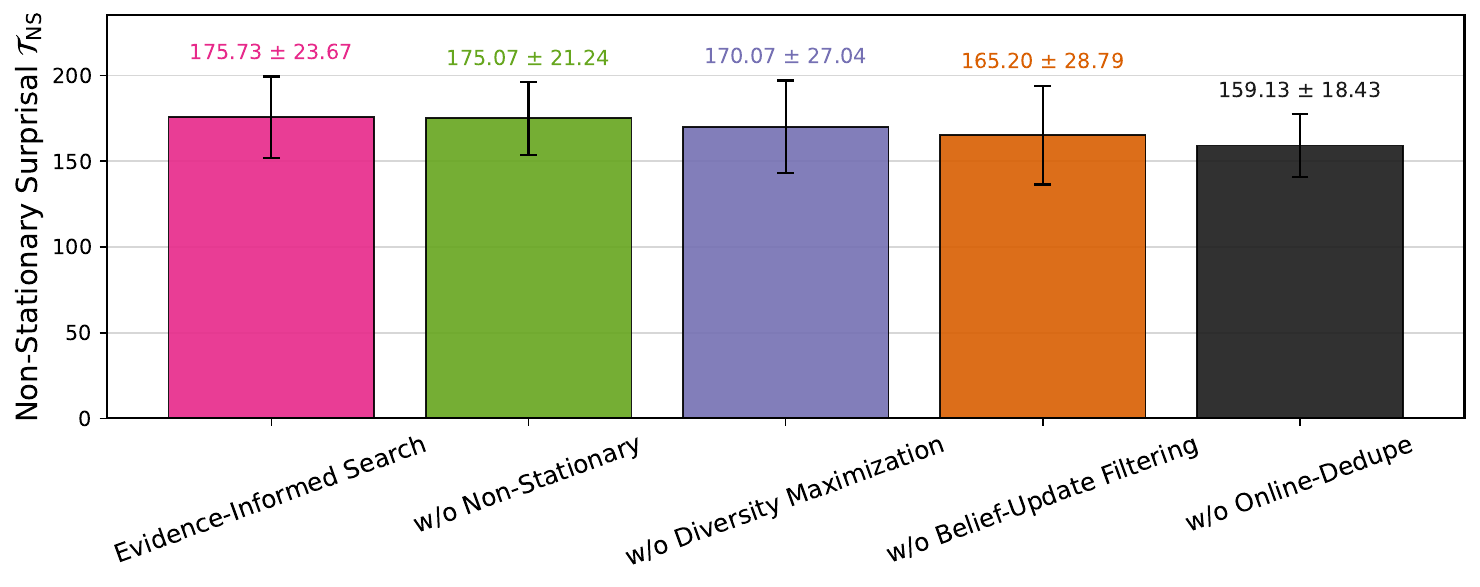}
        \caption{Ablation showing the effect of each component in evidence-informed search.}
        \label{fig:search-results-ablation-panel}
    \end{subfigure}

    \caption{
        \textbf{Evidence-informed search improves non-stationary discovery.}
        Search performance over 5 datasets and 3 repeat runs with $n=500$ experiments.
        \textbf{(a)} Evidence-informed search, which combines belief-update filtering with diversity maximization, outperforms standard \autodv{} search across reward and deduplication variants. Our method with non-stationary beliefs and online de-duplication performs best overall, yielding a 30.63\% gain ($\approx41$ surprisals) over standard static search and a 13.37\% gain ($\approx21$ surprisals) over standard search with online de-duplication.
        Online de-duplication consistently improves performance across methods, while simply replacing static surprisal rewards with non-stationary surprisal has an inconsistent effect.
        \textbf{(b)} Ablations on the evidence-informed search components show that the full method performs best, and switching to static rewards shows similar performance. Removing each of the other components---diversity maximization, belief-update filtering, and online-deduplication--- degrades performance (from lowest to highest effect), validating the complementary strengths of each of our modifications.
    }
    \label{fig:search-results}
\end{figure}

\paragraph{Setup.} We use the same evaluation setting as \S\ref{sec:static-to-non-stationary} and run experiments across five discovery domains from DiscoveryBench and BLADE using 3 repeat runs, each with a search budget of $n=500$ experiments. We use GPT-5-mini as the belief model and GPT-4o as the discovery agent. We measure performance using accumulated \emph{non-stationary surprisal}, computed with $\mathcal{C}_{\text{top-25}}$ as the evidence-informed context representation and GPT-5-mini as the belief model.

\paragraph{Methods compared.}
We evaluate our proposed search modifications against a set of ablations to isolate the contribution of each component.
As a baseline, we include the standard \autodv{} search. We then add each hypothesis selection mechanism, belief-update filtering and diversity maximization, independently. Finally, our full method combines belief-update filtering and diversity maximization. For each method, we also run variants with static as well as non-stationary rewards. Additionally, we also run each variant with and without online-deduplication.

\paragraph{Main results.}
Figure~\ref{fig:search-results}(a) compares standard \autodv{} search against our proposed evidence-informed search under accumulated non-stationary surprisal $\mathcal{T}_\text{NS}$. 
Evidence-informed search, which combines belief-update filtering with diversity maximization, consistently outperforms standard search across reward and de-duplication variants. 
The strongest variant uses non-stationary beliefs with online de-duplication, achieving a 30.63\% gain over standard static-reward search ($\approx 41$ additional surprisals) and a 13.37\% gain over standard search with online de-duplication ($\approx 21$ additional surprisals). 
This shows that the proposed hypothesis-selection mechanisms provide gains beyond de-duplication alone. 
We also find that online de-duplication improves performance across methods, confirming that near-duplicate hypotheses are a persistent source of wasted budget. 
In contrast, simply replacing the static search reward with non-stationary rewards has an inconsistent effect, reinforcing the need for explicit mechanisms to avoid redundancy and encourage diversity. In Figure~\ref{fig:search-results-all}, we also show the cumulative surprisal over the full search trajectory for each search variant.

\begin{wrapfigure}{r}{0.48\linewidth}
    \vspace{-1.5em}
    \centering
    \includegraphics[width=\linewidth]{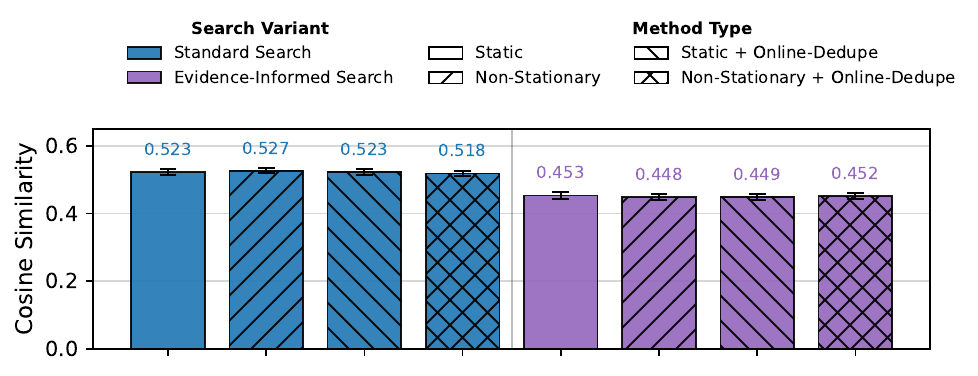}
    \caption{
    \textbf{Evidence-informed search improves semantic diversity.}
    Our method produces hypotheses with lower average pairwise cosine similarity than standard search, indicating broader semantic coverage. Switching from static to non-stationary beliefs alone does not change semantic diversity.
    }
    \label{fig:search-diversity-cosine}
    \vspace{-1.0em}
\end{wrapfigure}

\paragraph{Analyzing diversity.} 
Next, we analyze the semantic diversity of hypotheses found by each search variant\footnote{See the full plot in Figure~\ref{fig:search-diversity}.}. First, we see that online de-duplication increases the count of unique hypotheses across both standard and evidence-informed variants, as expected. However, as shown in Figure~\ref{fig:search-diversity-cosine}, evidence-informed search also produces hypotheses with greater semantic diversity, reflected in \emph{lower} average pairwise cosine similarity than standard search. By contrast, switching from static to non-stationary beliefs alone does not substantially change diversity, suggesting that the broader coverage comes primarily from the experiment-selection modifications we introduce in evidence-informed search rather than the choice of beliefs.

\paragraph{Ablations.} The ablations in Figure~\ref{fig:search-results}(b) show that all components of evidence-informed search, except the choice of surprisal reward, contribute to performance. The full method performs best, while removing diversity maximization, belief-update filtering, and online de-duplication each degrades cumulative surprisal, showing that each of our introduced modifications provides complementary gains by respectively avoiding already-implied hypotheses, encouraging broader coverage, and suppressing redundant discoveries.

\section{Related Work}
\label{sec:related}
\paragraph{LLM-based scientific discovery.}

Most existing AI-scientists treat discovery as an episodic problem. When applied in an open-ended setting, they typically expect a dataset, objective, literature context, or a prompt, and optimize for an output conditioned on the fixed input \citep{Bran2023AugmentingLL, majumder2024position}. While AI-scientist \citep{yamada2025aiscientistv2workshoplevelautomated, aiscientist} uses search-trees for optimized agentic actions, such as managing experiments, refining figures, and producing full manuscripts, it does not use intermediate results to update the frame of reference for novelty detection. Google AI co-scientist\footnote{\url{https://research.google/blog/accelerating-scientific-breakthroughs-with-an-ai-co-scientist/}} operates on a goal-driven setup, where it formulates hypotheses based on the literature related to the given objective; however, it does not accumulate evidence to update its cumulative world model, which could be used to inform newer sub-discoveries. Finally, Kosmos \citep{mitchener2025kosmos}, while working in a goal-driven setup, internally develops a world model to inform literature search or the next data analysis task. While Kosmos presents the closest system that uses intermediate findings to inform subsequent hypothesis generation, it does not use these findings to update inference over the world model to reflect a non-stationary belief. %
Evidence-informed next action selection approaches akin to those presented here could be used to similarly augment these systems for their native objectives and to elicit non-stationary beliefs representing the cumulative scientific discovery knowledge uncovered.

\paragraph{Non-parametric LLM memory.} 
Recent work on memory-augmented LLMs has explored external memory mechanisms for coherent, consistent, and efficient reasoning. These systems may provide memory capabilities that merely augment an LLM's reasoning capabilities via context manipulation or provide their own native reasoning capabilities. BeliefBank \citep{kassner-etal-2021-beliefbank} treats the language model as a subcomponent as part of a larger system, which maintains a symbolic memory of beliefs and constraint-based reasoning to improve language model belief consistency without requiring retraining the underlying model. Dealing with the finite nature of LLM context windows presents an additional challenge in determining which context to provide to the LLM, and how to do so efficiently. InfLLM~\citep{xiao2024infllm} explores a training-free method that stores distant context, which can then be efficiently retrieved based on their relevance to enable effective processing of extremely long sequences. Finding ideal ways to store, organize, and consolidate information is also an active area of research that includes tree~\citep{rezazadeh2025from}, graph~\citep{chhikara2025mem0}, and operating systems like hierarchical-based~\citep{kang-etal-2025-memory} structures.

\paragraph{Parametric Memory.}
Recent work on parametric and semi-parametric memory studies how models can internalize information beyond the transient prompt, either by writing new information into compact trainable states or by retrieving parameter-like updates at inference time. GradMem learns to compress a context into a small set of writable prefix-memory tokens through test-time gradient descent, providing a loss-driven mechanism for storing contextual information while keeping the base model fixed \citep{kuratov2026gradmem}. Similarly, MIRA augments a shared backbone with Hopfield-style associative memories that retrieve adapter-weight updates, enabling per-sample modulation across task and domain shifts \citep{agrawal2025mira}. These approaches are closely related to our goal of making models adapt to newly observed evidence, but they primarily treat memory as a mechanism for storing or retrieving task-relevant information. In contrast, our work focuses on the epistemic role of memory: evidence is not merely retained, but used to update an LLM's in-context beliefs so that future hypotheses are evaluated relative to a non-stationary belief state.

\paragraph{Test-time adaptation and LLM-based Bayesian optimization.}
A growing body of work treats inference not as a one-shot prediction problem, but as an adaptive process in which model behavior changes in response to test-time feedback. Test-time training methods update model parameters using self-supervised objectives at inference time to improve robustness under distribution shift \citep{sun2020test}; recent LLM-oriented variants extend this idea to language models by adapting them to unlabeled test inputs or by optimizing policies online \citep{hu2025test,phan2025migrate}. Other work adapts LLM outputs without parameter updates through iterative feedback and refinement, as in Self-Refine \citep{madaan2023selfrefine}. A complementary line uses LLMs as black-box optimizers, where previously evaluated candidates and their rewards are placed in context to guide future proposals. OPRO casts optimization itself as prompting, using an LLM to iteratively generate improved solutions conditioned on past solution--score pairs \citep{yang2024large}, while LLAMBO integrates LLMs into Bayesian optimization for warm-starting, surrogate modeling, and candidate generation \citep{liu2024large}. More directly, Bayesian-OPRO combines LLM proposal generation with Bayesian optimization to search over candidate solutions using evolving uncertainty estimates \citep{agarwal2025searching}. Our work shares this sequential, evidence-conditioned view of inference, but shifts the object of adaptation from parameters, prompts, or candidate solutions to the model's \emph{belief state} in context. By explicitly updating LLM beliefs with observed evidence, our method induces a non-stationary discovery objective: evidence that was initially surprising should cease to be surprising once incorporated, allowing search to prioritize hypotheses that remain informative under the current epistemic state.

\section{Limitations}
\label{sec:limitations}

First, our belief estimates are elicited from LLM samples and therefore depend on prompting choices, sampling variance, calibration quality, and the model's ability to faithfully express uncertainty. 
Second, while our non-stationary objective reduces redundant rediscovery, it may also suppress hypotheses that are legitimately worth revisiting under stronger or conflicting evidence. 
Third, our experiments focus on controlled discovery settings with finite search budgets; scaling the method to open-ended scientific workflows will require better evidence verification, stronger uncertainty calibration, and mechanisms for long-term memory beyond the current context window.

\section{Conclusion}
\label{sec:conclusion}

We show that surprisal in continual LLM-driven discovery should be non-stationary: hypotheses must be evaluated relative to beliefs updated by prior evidence. Static surprisal, as used in \autodv{}, can reward hypotheses already implied by earlier discoveries, leading to redundant search. We address this with evidence-informed LLM beliefs, updating priors in context from previously tested hypotheses and using the resulting belief state to compute non-stationary surprisal. We find that embedding-based retrieval over prior discoveries best approximates posterior beliefs, revealing that many static surprisals are spurious. Finally, we incorporate non-stationary beliefs into search through belief-update filtering and diversity maximization, improving accumulated non-stationary surprisal across five discovery domains. These results suggest that long-horizon LLM discovery systems must not only generate and verify hypotheses, but also continually revise what counts as surprising.

\bibliographystyle{abbrvnat}
\bibliography{custom}

\newpage
\appendix

\section{Additional Analyses}
\label{app:additional-analyses}

\subsection[Context Representations: Top-k Comparisons]{Context Representations: Top-$k$ Comparisons}
\begin{figure}[h]
    \centering
    \includegraphics[width=\textwidth]{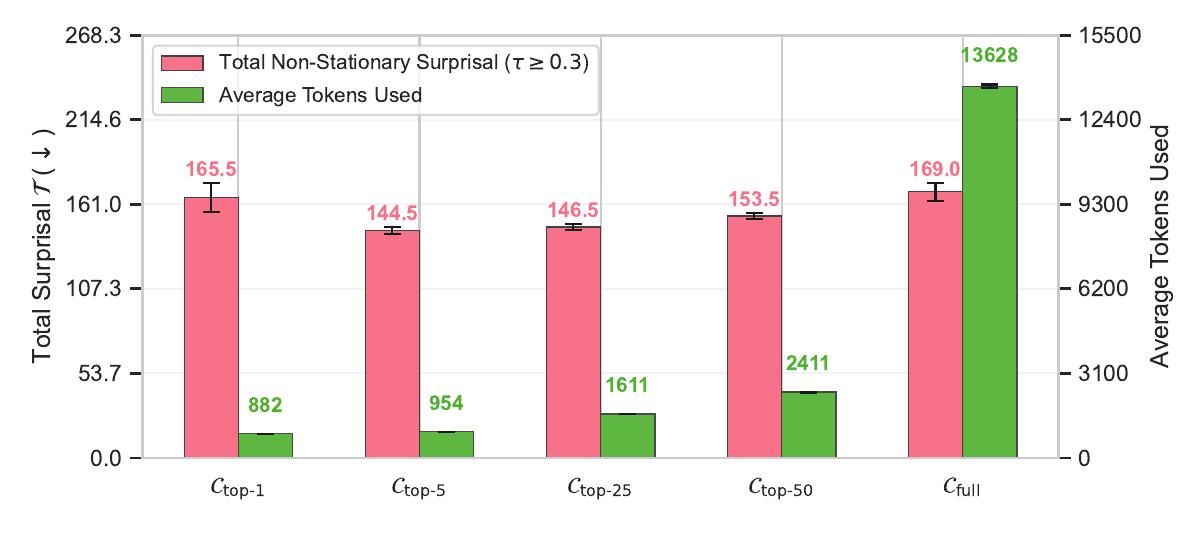}
    \caption{
        \textbf{Effect of $k$ in $\mathcal{C}_{\text{top-}k}$.}
        We vary the number of retrieved prior discoveries used to construct the evidence-informed context and report both the total number of surprisals in the run and the average number of input+output tokens used.
        We find that increasing $k$ does not monotonically improve performance, instead showing best performance at $k=5$ and $k=25$, while saving $14.34\times$ and $8.45\times$ fewer tokens, respectively, than using the full search trajectory.
    }
    \label{fig:spurious-surprisal-long-horizon}
\end{figure}

\subsection{Context Representations: Reasoning Effort}
\begin{figure}[h]
    \centering
    \includegraphics[width=\textwidth]{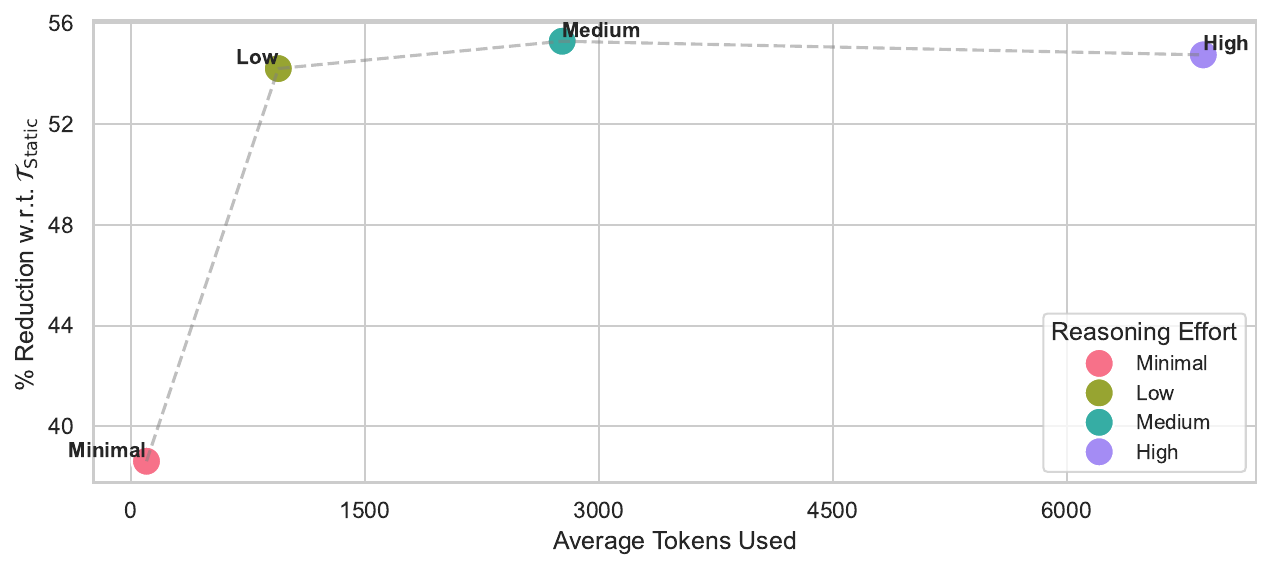}
    \caption{
        \textbf{Performance-cost tradeoff at different levels of LLM reasoning.}
        We vary the GPT-5-mini reasoning effort setting from \texttt{minimal} to \texttt{high} and plot the reduction in total surprisals as compared to a static beliefs run against the average number of input+output tokens used.
        Higher reasoning effort can improve surprisal reduction but increases inference cost, allowing us to select the \texttt{low} setting, which is at the ``elbow'' (point of diminishing returns). Surprisingly, we also find that surprisal reduction diminishes slightly at the highest reasoning setting.
    }
    \label{fig:spurious-surprisal-long-horizon-cost}
\end{figure}

\newpage

\subsection{Improving Search: Belief-Update Filtering}
\begin{figure}[h]
    \centering
    \includegraphics[width=\textwidth]{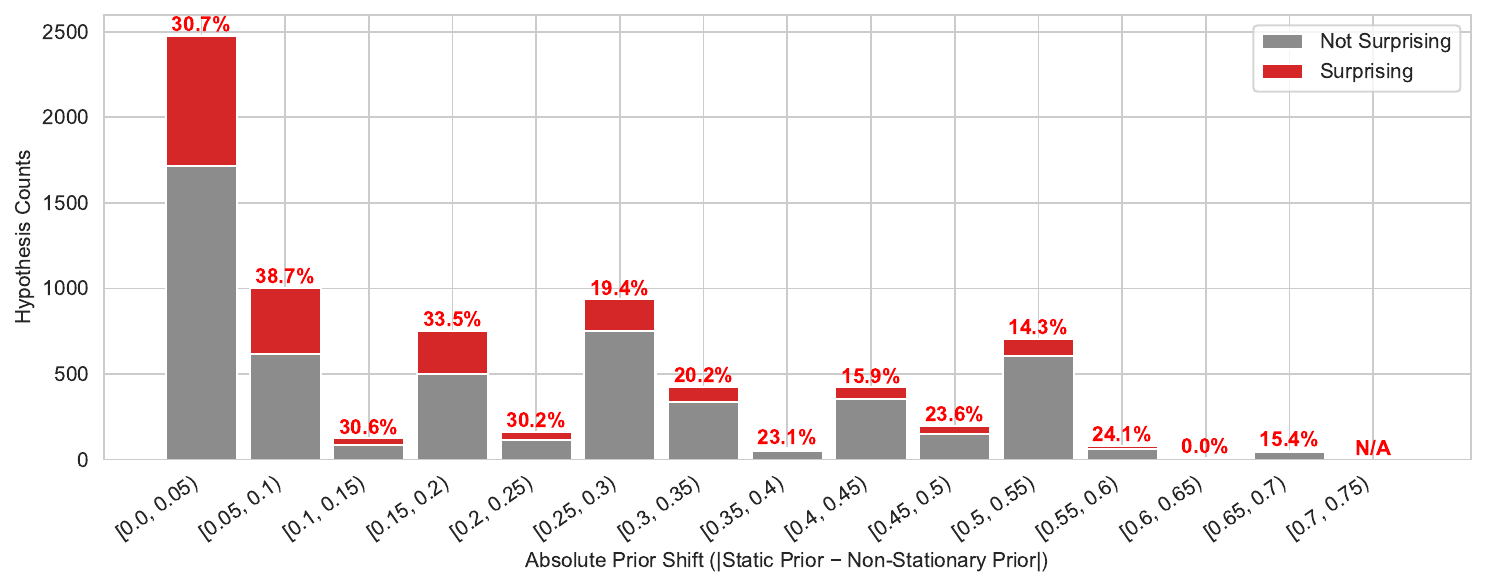}
    \caption{
        \textbf{Proportion of non-stationary surprisals as a function of prior belief shift.}
        Across hypotheses generated from 5 datasets and 3 repeat runs using static search, we find that there is a decreasing trend, which emerges at a shift threshold of 0.2, where as belief shift increases, the proportion of non-stationary surprisals goes down. This indicates that when beliefs significantly update with evidence from past discoveries, the present hypotheses may often be derivable.
    }
    \label{fig:belief-update-filtering-analysis}
\end{figure}

\subsection{Improving Search: All Results}
\begin{figure}[h]
    \centering
    \includegraphics[width=\textwidth]{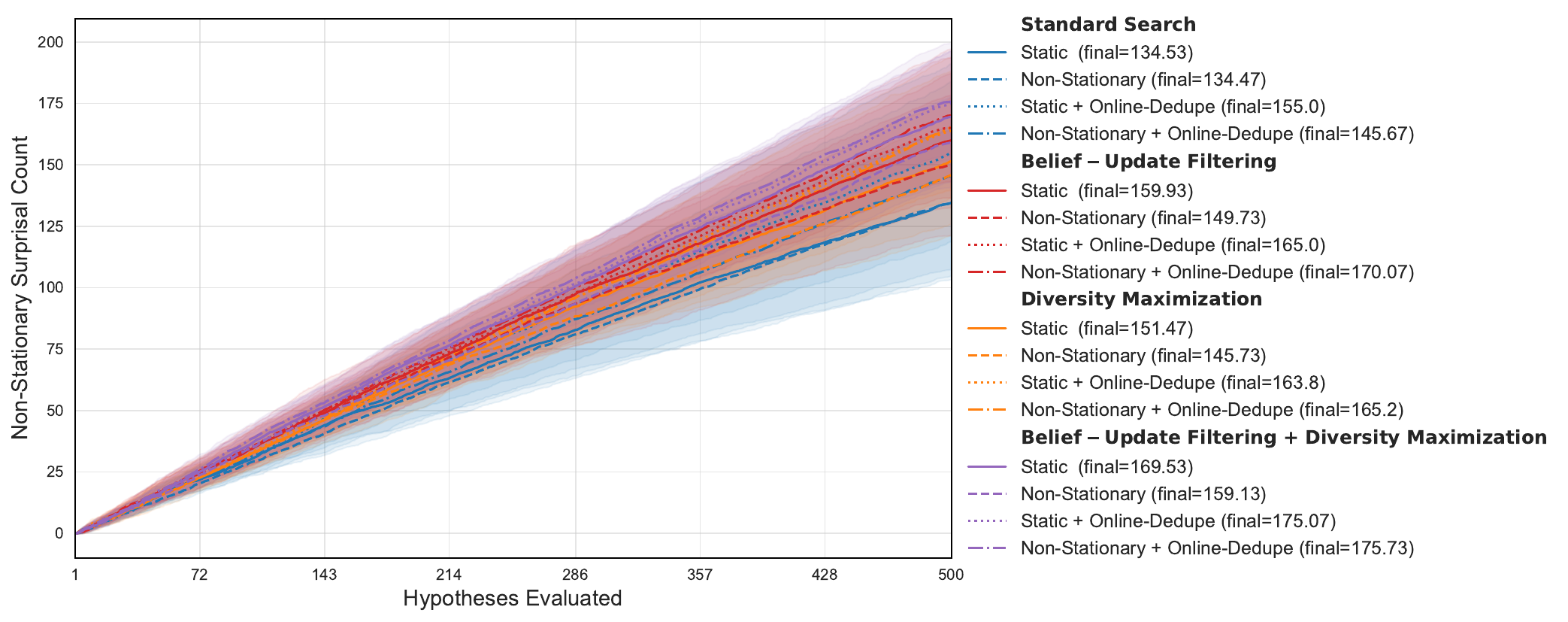}
    \caption{
        \textbf{Search improvements under non-stationary surprisal evaluation.} Search variants by accumulated non-stationary surprisal count across 5 discovery domains and 3 repeats runs. 
        Our proposed method combining belief-update filtering with diversity maximization (with online-deduplication) shows the best performance across all methods, yielding a 30.63\% gain ($\approx41$ surprisals) over the original static-reward search and a 13.37\% gain ($\approx21$ surprisals) over original search + online-deduplication, showing a 17.26 percentage point contribution from our hypothesis selection modifications.
        Online de-duplication, additionally, consistently improves performance across all methods, while simply replacing static surprisal rewards with non-stationary surprisal has no consistent effect. 
    }
    \label{fig:search-results-all}
\end{figure}

\subsection{Improving Search: Hypothesis Diversity}
\begin{figure}[h]
    \centering
    \includegraphics[width=\linewidth]{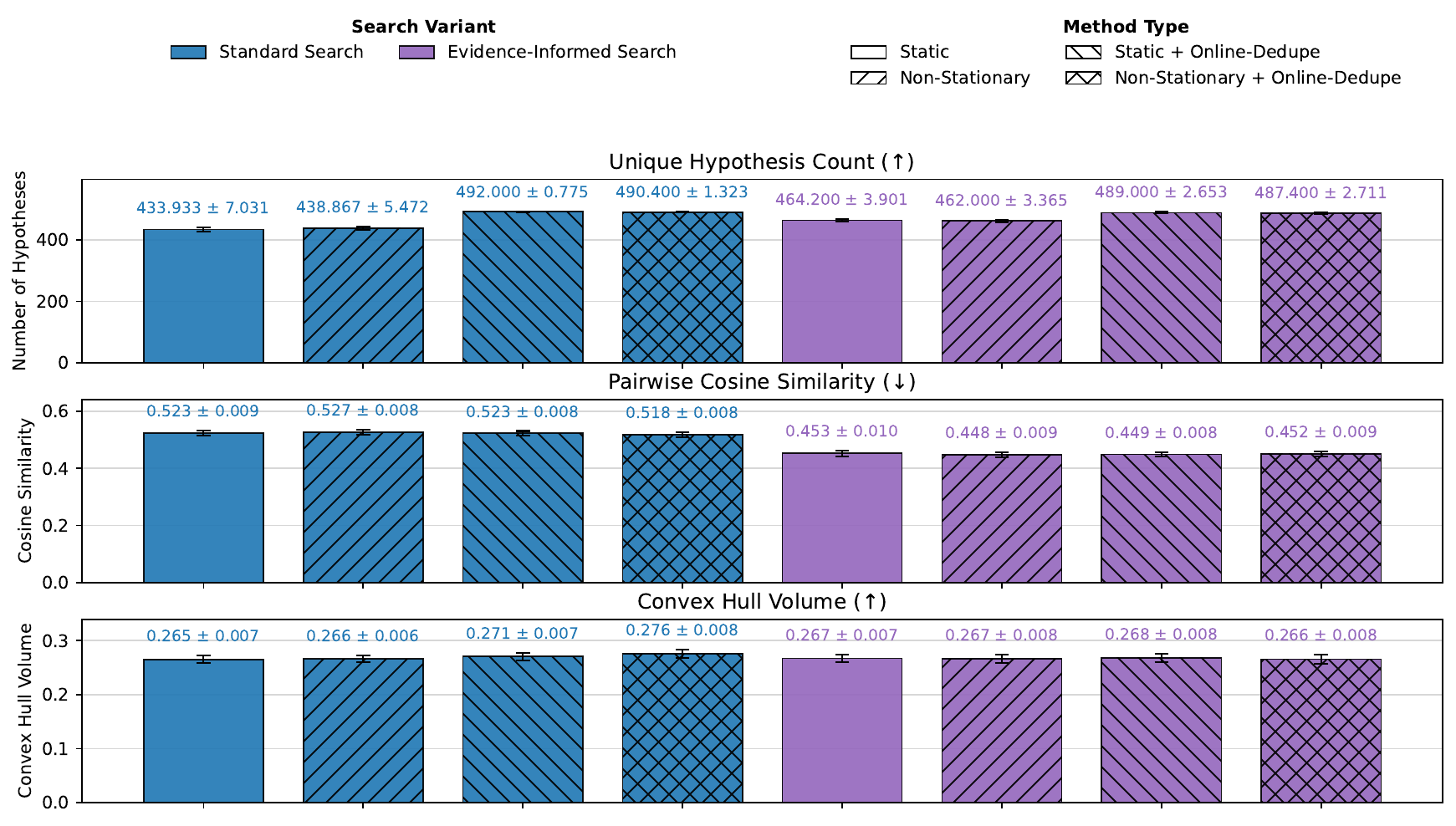}
    \caption{
    \textbf{Evidence-informed search improves semantic diversity.}
    We compare the diversity of hypotheses produced by standard search and Evidence-Informed Search across static and non-stationary belief variants, with and without online de-duplication.
    As expected, adding online de-duplication increases the number of unique hypotheses across variants.
    Uniqueness alone does not capture semantic diversity: hypotheses discovered by Evidence-Informed Search exhibit lower average pairwise cosine similarity than those found by standard search, indicating broader semantic coverage.
    In contrast, replacing static beliefs with non-stationary beliefs by itself does not induce a meaningful diversity difference, suggesting that the diversity gains arise primarily from the experiment selection mechanisms we introduce in evidence-informed search rather than the belief representations. Lastly, we find that the PCA-reduced 3-dimensional convex hull volume is unable to provide meaningful discrimination between the different methods.
    }
    \label{fig:search-diversity}
\end{figure}

\section{Human Expert Study}
\label{app:human-study}
TBA.

\section{LLM Prompts}
\label{app:prompts}

\subsection{Non-Stationary Belief Elicitation}
\label{app:belief-elicitation-prompt}

\begin{promptbox}{System Prompt}
You are a research scientist skilled at analyzing scientific hypotheses.
Your task is to provide your belief about the given hypothesis.
Use your prior knowledge and any previously tested hypotheses
as potential evidence to support or refute the hypothesis.
Trust the evidence when it is relevant;
trust your prior knowledge when it is more convincing.
\end{promptbox}

\begin{promptbox}{User Prompt}
Hypothesis: \{hypothesis\}

=== Previous Hypotheses ===

1. \{previous hypothesis 1\} (belief based on experimental evidence: {p1})

2. \{previous hypothesis 2\} (belief based on experimental evidence: {p2})

3. \{previous hypothesis 3\} (belief based on experimental evidence: {p3})
\(\cdots\)

Here's the hypothesis again: \{hypothesis\}

Reason carefully before making your assessment.
\end{promptbox}

\newpage
\section*{NeurIPS Paper Checklist}

\begin{enumerate}

\item {\bf Claims}
    \item[] Question: Do the main claims made in the abstract and introduction accurately reflect the paper's contributions and scope?
    \item[] Answer: \answerYes{} %
    \item[] Justification: We have reviewed the literature sufficiently and run sufficient experiments to be confident in our claims.

\item {\bf Limitations}
    \item[] Question: Does the paper discuss the limitations of the work performed by the authors?
    \item[] Answer: \answerYes{}{} %
    \item[] Justification: Please see Section 6 Limitations.

\item {\bf Theory assumptions and proofs}
    \item[] Question: For each theoretical result, does the paper provide the full set of assumptions and a complete (and correct) proof?
    \item[] Answer: \answerNA{} %
    \item[] Justification: The paper does not include theoretical results.

    \item {\bf Experimental result reproducibility}
    \item[] Question: Does the paper fully disclose all the information needed to reproduce the main experimental results of the paper to the extent that it affects the main claims and/or conclusions of the paper (regardless of whether the code and data are provided or not)?
    \item[] Answer: \answerYes{} %
    \item[] Justification: The paper includes details where necessary. We will additionally be open-sourcing all our code.

\item {\bf Open access to data and code}
    \item[] Question: Does the paper provide open access to the data and code, with sufficient instructions to faithfully reproduce the main experimental results, as described in supplemental material?
    \item[] Answer: \answerNo{} %
    \item[] Justification: We do not yet include access to the code in the paper, but commit to releasing it.

\item {\bf Experimental setting/details}
    \item[] Question: Does the paper specify all the training and test details (e.g., data splits, hyperparameters, how they were chosen, type of optimizer) necessary to understand the results?
    \item[] Answer: \answerYes{} %
    \item[] Justification: We include all details whenever experiments setup was required.

\item {\bf Experiment statistical significance}
    \item[] Question: Does the paper report error bars suitably and correctly defined or other appropriate information about the statistical significance of the experiments?
    \item[] Answer: \answerYes{} %
    \item[] Justification: Yes, we report std. and error bars in all our results.

\item {\bf Experiments compute resources}
    \item[] Question: For each experiment, does the paper provide sufficient information on the computer resources (type of compute workers, memory, time of execution) needed to reproduce the experiments?
    \item[] Answer: \answerYes{} %
    \item[] Justification: All our experiments were run on CPU with LLM API keys, so we do not need heavy computing machinery. Our runs do take long, which we will further clarify in rebuttal and later.
    
\item {\bf Code of ethics}
    \item[] Question: Does the research conducted in the paper conform, in every respect, with the NeurIPS Code of Ethics \url{https://neurips.cc/public/EthicsGuidelines}?
    \item[] Answer: \answerYes{} %
    \item[] Justification: Yes, it does.

\item {\bf Broader impacts}
    \item[] Question: Does the paper discuss both potential positive societal impacts and negative societal impacts of the work performed?
    \item[] Answer: \answerYes{} %
    \item[] Justification: We discuss the positive impact our work could have on long-horizon discovery as well as on other discovery systems. We have also talked about limitations.
    
\item {\bf Safeguards}
    \item[] Question: Does the paper describe safeguards that have been put in place for responsible release of data or models that have a high risk for misuse (e.g., pre-trained language models, image generators, or scraped datasets)?
    \item[] Answer: \answerNA{} %
    \item[] Justification: Our paper poses no such risk.

\item {\bf Licenses for existing assets}
    \item[] Question: Are the creators or original owners of assets (e.g., code, data, models), used in the paper, properly credited and are the license and terms of use explicitly mentioned and properly respected?
    \item[] Answer: \answerYes{} %
    \item[] Justification: Yes, we have credited authors appropriately.

\item {\bf New assets}
    \item[] Question: Are new assets introduced in the paper well documented and is the documentation provided alongside the assets?
    \item[] Answer: \answerNA{} %
    \item[] Justification: No new assets have been released.

\item {\bf Crowdsourcing and research with human subjects}
    \item[] Question: For crowdsourcing experiments and research with human subjects, does the paper include the full text of instructions given to participants and screenshots, if applicable, as well as details about compensation (if any)? 
    \item[] Answer: \answerYes{} %
    \item[] Justification: This information will be provided in the supplemental.

\item {\bf Institutional review board (IRB) approvals or equivalent for research with human subjects}
    \item[] Question: Does the paper describe potential risks incurred by study participants, whether such risks were disclosed to the subjects, and whether Institutional Review Board (IRB) approvals (or an equivalent approval/review based on the requirements of your country or institution) were obtained?
    \item[] Answer: \answerNA{} %
    \item[] Justification: Our paper does not require IRB approval.

\item {\bf Declaration of LLM usage}
    \item[] Question: Does the paper describe the usage of LLMs if it is an important, original, or non-standard component of the core methods in this research? Note that if the LLM is used only for writing, editing, or formatting purposes and does \emph{not} impact the core methodology, scientific rigor, or originality of the research, declaration is not required.
    \item[] Answer: \answerNA{} %
    \item[] Justification: The method development in this paper was not done by LLMs.

\end{enumerate}

\end{document}